\documentclass[final]{sabra}

\usepackage{times}
\usepackage{epsfig}
\usepackage{graphicx}
\usepackage{amsmath}
\usepackage{amssymb}
\usepackage{booktabs} %
\usepackage{soul}
\usepackage{xcolor}

\usepackage{multirow}

\usepackage[pagebackref=true,breaklinks=true,colorlinks,bookmarks=false]{hyperref}

\makeatletter 
\newcommand\figcaption{\def\@captype{figure}\caption} 
\newcommand\tabcaption{\def\@captype{table}\caption} 
\makeatother

\newcommand{\subobj}{[\emph{subject}, \emph{object}] }
\newcommand{\subobjrel}{[\emph{subject}, \emph{predicate}, \emph{object}] }
\newcommand{\fbbox}{f_\textrm{box}}
\newcommand{\frel}{f_\textrm{rel}}
\newcommand{\sneg}{S_\textrm{neg}}
\newcommand{\spos}{S_\textrm{pos}}

\title{Towards Overcoming False Positives in Visual Relationship Detection}
\author{Daisheng Jin\textsuperscript{\rm 1,2}\thanks{Equal contribution}, Xiao Ma\textsuperscript{\rm 3}\footnotemark[1], Chongzhi Zhang\textsuperscript{\rm 1,2}\footnotemark[1], Yizhuo Zhou\textsuperscript{\rm 4}, Jiashu Tao\textsuperscript{\rm 3}, \\
Mingyuan Zhang\textsuperscript{\rm 1}\thanks{Corresponding author, zhangmingyuan@sensetime.com}, Haiyu Zhao\textsuperscript{\rm 1}, Shuai Yi\textsuperscript{\rm 1}, Zhoujun Li\textsuperscript{\rm 2}, Xianglong Liu\textsuperscript{\rm 2}, Hongsheng Li\textsuperscript{\rm 5} \\
\\
\textsuperscript{\rm 1} SenseTime Research, \textsuperscript{\rm 2} Beihang University, \textsuperscript{\rm 3} National University of Singapore \\
\textsuperscript{\rm 4} ByteDance AI Lab, \textsuperscript{\rm 5} Multimedia Laboratory, The Chinese University of Hong Kong \\
}

\def\cvprPaperID{5048} %
\def\confYear{CVPR 2021}

\begin{document}
\maketitle

\begin{abstract}

In this paper, we investigate the cause of the high false positive rate in Visual Relationship Detection (VRD). We observe that during training, the relationship proposal distribution is highly imbalanced: most of the negative relationship proposals are easy to identify, e.g., the inaccurate object detection, which leads to the under-fitting of low-frequency difficult proposals. This paper presents \textit{Spatially-Aware Balanced negative pRoposal sAmpling} (SABRA), a robust VRD framework that alleviates the influence of false positives. 
To effectively optimize the model under imbalanced distribution,
SABRA adopts \textit{Balanced Negative Proposal Sampling} (BNPS) strategy for mini-batch sampling. BNPS divides proposals into 5 well defined sub-classes and generates a balanced training distribution according to the inverse frequency. BNPS gives an easier optimization landscape and significantly reduces the number of false positives. To further resolve the low-frequency challenging false positive proposals with high spatial ambiguity, we improve the spatial modeling ability of SABRA on two aspects:
a simple and efficient multi-head heterogeneous graph attention network (MH-GAT) that models the global spatial interactions of objects, and a spatial mask decoder that learns the local spatial configuration.
SABRA outperforms SOTA methods by a large margin on two human-object interaction (HOI) datasets and one general VRD dataset.

\end{abstract}

\section{Introduction}
Visual Relationship Detection (VRD) is an important visual task that bridges the gap between middle-level visual perception, e.g., object detection, and high-level visual understanding, e.g., image captioning~\cite{vinyals2016show,gan2017stylenet}, and visual question answering~\cite{lu2016hierarchical}.   General VRD aims to understand the interaction between two arbitrary objects in the scene. Human-Object Interaction (HOI), as a specific case of VRD, focuses on understanding the interaction between humans and objects, e.g., woman-cut-cake. 

Existing VRD methods focus on building powerful feature extractors for each \subobj pair, predicting the \emph{predicate} between the \emph{subject} and the \emph{object}, and outputting \subobjrel triplet predictions. Some prior works model subject and object relationship independently~\cite{galleguillos2008object,ramanathan2015learning}, which loses the global context and is susceptible to inaccurate detections. Some recent works incorporate the union bounding boxes of the \subobj as additional features to provide additional spatial information~\cite{DBLP:conf/eccv/QiWJSZ18,DBLP:conf/iccv/WanZLLH19} or use the graph neural networks (GNNs) to better extract global object relationships~\cite{yao2018exploring,hu2019neural,mi2020hierarchical}. However, for any visual relationship detector, 90\% of its predictions are false positives because according to our statistical result, an image normally contains less than 10 positive relationships, while detectors consider the top-$100$ predictions.

\begin{figure*}[t]
    \centering
    \begin{tabular}{c c}
        \includegraphics[width=0.25\linewidth]{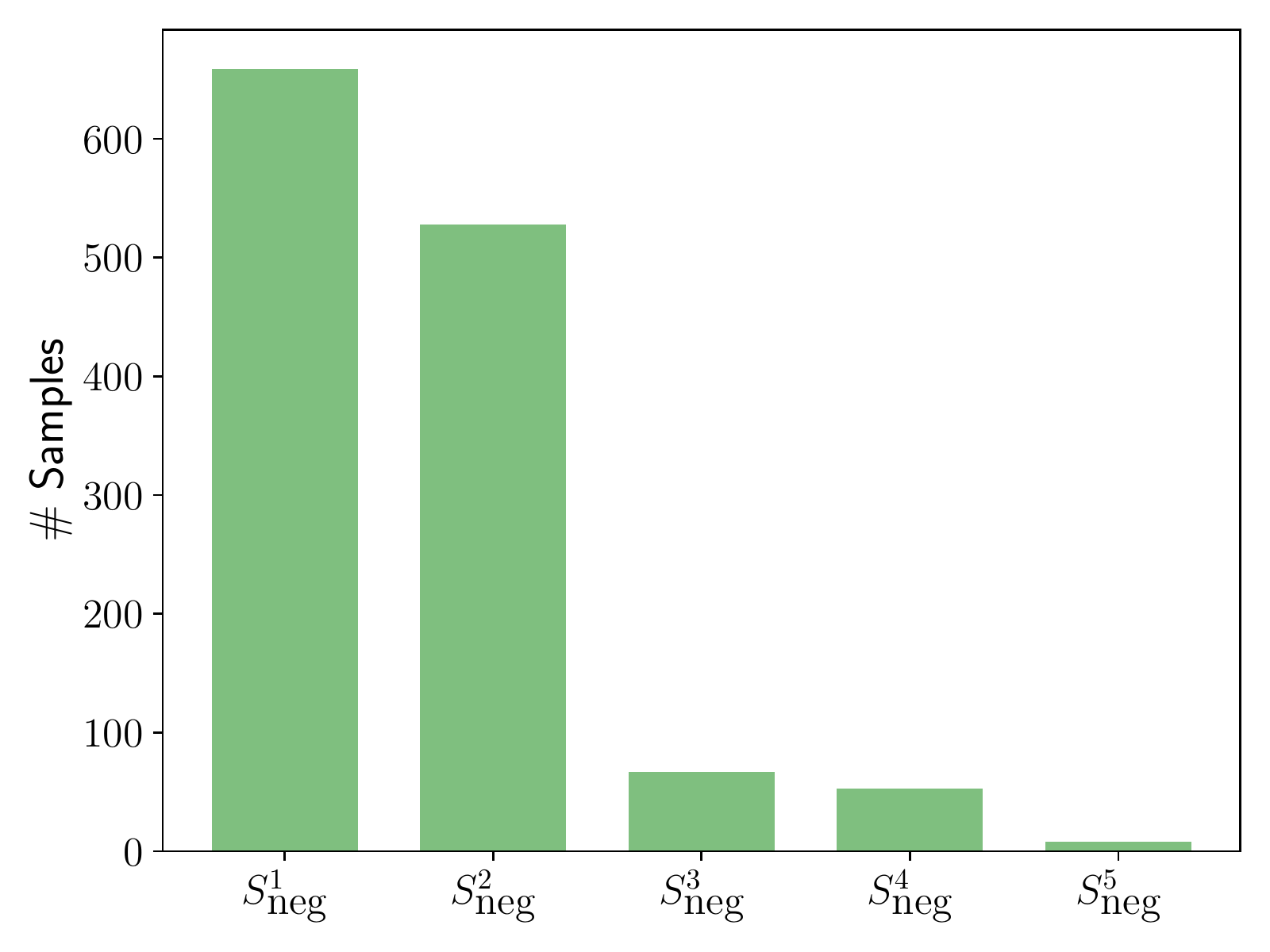}&
        \includegraphics[width=0.45\linewidth]{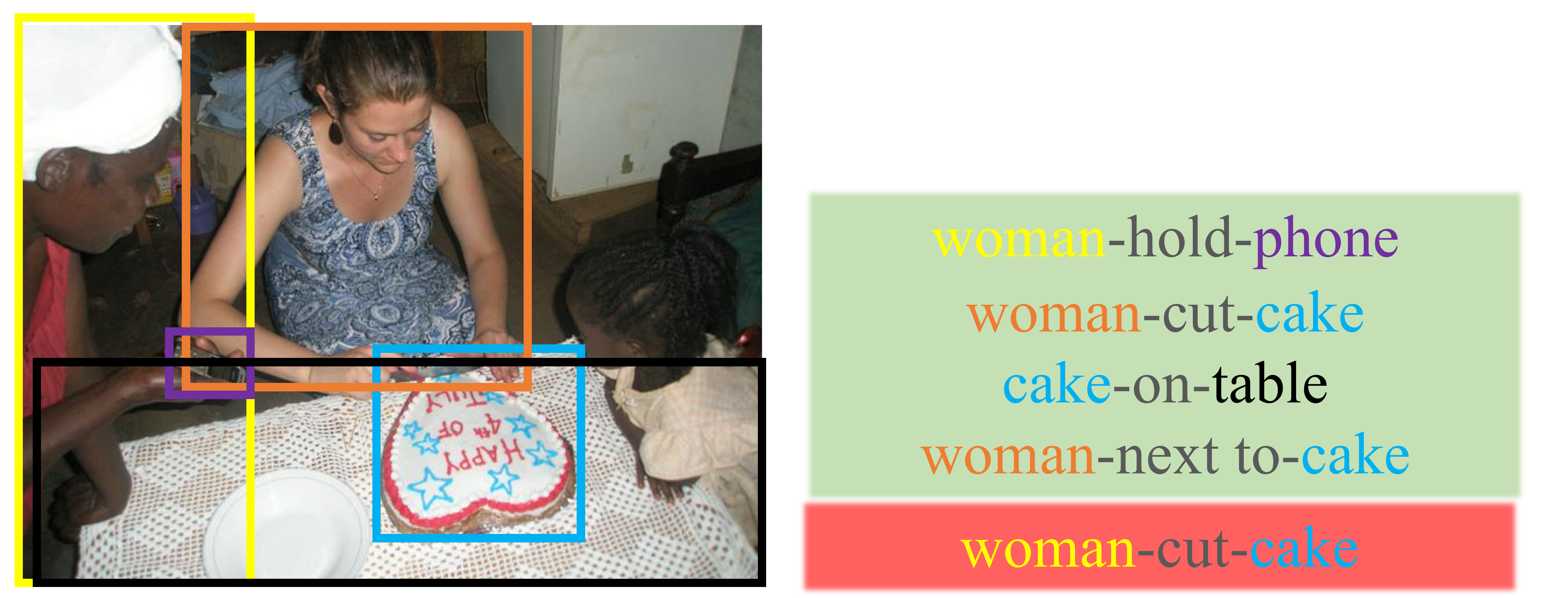}\\
        (a) Imbalanced negative proposal distribution & (b) Influence of ambiguous context 
    \end{tabular}
    \centering
    \caption{(a) VRD normally gives an extremely imbalanced negative proposal distribution, where $\sneg^{1:5}$ denotes 5 different types of negative proposals. Detailed definition of $\sneg^{1:5}$ is given by Eqn.~\ref{eqn:bnps}. (b) Contextual objects introduce ambiguous information to the relationship classification, e.g., the model thinks the woman on the left is cutting the cake by mistake.
    }
    \label{fig:errors}
\end{figure*}

In addition to the high spatial ambiguity that prior works have been focusing on, we observe that another critical reason for the high false positive rate in VRD is the difficult optimization landscape caused by extremely imbalanced relationship proposal distribution~\cite{ren2020balanced}. 
A typical visual relationship detector takes as input a set of \subobj proposals. However, most of them are negative proposals, i.e., no relationship exists for given \subobj pairs.
We observe that the negative proposals are introduced mainly by two sources with different levels of difficulties: inaccurate single object detections and the incorrect \subobj associations. Inaccurate detections, which contribute to more than 90\% of the negative proposals (first two columns in Fig.~\ref{fig:errors}.a), can be easily identified by the appearance feature of the single object detections. Incorrect associations contribute to less than 10\% of negative proposals (last three columns in Fig.~\ref{fig:errors}.a) but are significantly more difficult: the subject might be related to other objects, but the selected \subobj pair can be unrelated, which requires careful inspection of the global context. 

We present \textit{Spatially-Aware Balanced negative pRoposal sAmpling} (SABRA), a robust and general VRD framework that alleviates the influence of false positives for both HOI and general VRD tasks. 
According to the two sources of negative proposals, \textit{i.e.}, inaccurate detections, and incorrect associations, we introduce a division of negative proposals into 5 sub-classes $\sneg^{1:5}$. For each single object detection, we consider (1) if it is an accurate detection, and (2) if it is in any relationship with a different object. Specifically, $\sneg^{1:2}$ cover the inaccurate detections, and $\sneg^{3:5}$ discuss the incorrect associations. From $\sneg^1$ to $\sneg^5$, the sample size decreases, and the classification difficulty increases, because detecting the false positives according to the accuracy of detection is no longer sufficient, and careful understanding about object relationships becomes necessary for the task.
Detailed definitions can be found in Eqn.~\ref{eqn:bnps}. As visualized in Fig.~\ref{fig:errors}.a, the sample sizes of 5 negative proposal sub-classes give a highly imbalanced distribution, which degrades the performance of data-driven VRD algorithms.
Inspired by the learning under imbalanced distribution literature~\cite{wang2017learning,byrd2019effect}, we alleviate the optimization difficulty by \textit{Balanced Negative Proposal Sampling} scheme. 
BNPS computes the statistics of each class and performs a simple yet effective \textit{Class Balanced Sampling}~\cite{DBLP:conf/eccv/ShenLH16} for balanced data distribution. 
Balanced negative proposal sampling significantly reduces the number of false positive occurrences.
For low-frequency difficult classes $\sneg^{3:5}$, e.g., Fig.~\ref{fig:errors}.b, we further improve the spatial modeling of SABRA on two aspects.
For the global context understanding, SABRA extends the existing GNN-based methods~\cite{xu2019learning,mi2020hierarchical} with a heterogeneous message passing scheme which effectively addresses the distribution divergence between different features. For local spatial configuration, SABRA learns a position-aware embedding vector by predicting the locations in each \subobj pair. 

In our experiment, we evaluate SABRA on two commonly used HOI datasets, V-COCO, and HICO-DET, and one general object relationship detection dataset, VRD. We show that SABRA significantly outperforms SOTA methods by a large margin. We also visualize the results and show that SABRA effectively reduces the false positives in VRD and misclassification in terms of the spatial ambiguity.

\section{Related Works}

\subsection{Visual Relationship Detection}
VRD is an important middle-level task bridging low-level visual recognition with high-level visual understanding. Earlier works focus on \textit{Support Vector Machines} (SVMs) with discriminative features for relationship classification~\cite{delaitre2010recognizing,delaitre2011learning}. However, handcrafting features is difficult and harmful to the final performance of VRD algorithms. With the advances of deep learning, data-driven approaches are widely adopted for VRD. Specifically, \textit{Convolutional Neural Networks} (CNNs) are used for automatic feature extraction and information fusion, which achieved great improvement in VRD~\cite{DBLP:conf/iccv/WanZLLH19,xu2019interact,wang2020learning}. \textit{Graph Neural Networks} (GNNs) further improve the feature extraction process by explicitly modeling the instance-wise interactions between objects~\cite{xu2019learning,mi2020hierarchical}. Due to the nature of VRD tasks, additional information has been introduced as auxiliary training signals, such as language priors~\cite{lu2016visual}, prior interactiveness knowledge of objects~\cite{li2019transferable}, and action co-occurrence knowledge~\cite{kim2020detecting}.
In comparison with the existing methods, SABRA is the first to identify the importance of false positives in VRD tasks and has significantly outperformed SOTA methods in our experiments.

\subsection{Learning under Imbalanced Distribution}
Real-world data is imbalanced by nature: a few high-frequency classes contribute to most of the samples, while a large number of low-frequency classes are under-represented in data. Standard imbalanced learning techniques include data re-balancing~\cite{barandela2003restricted,buda2018systematic,byrd2019effect}, loss function engineering~\cite{cao2019learning,hayat2019gaussian} and meta-learning~\cite{jamal2020rethinking}. VRD, as a common computer vision task, also suffers from the imbalanced problem~\cite{lin2020gps,kim2020detecting}. Specifically, \cite{kim2020detecting} considers the imbalance of relationship imbalance, i.e., the imbalance of positive samples, and uses action co-occurrence to provide additional labels.
However, none of the existing methods consider the imbalance of the negative \subobj proposals, which commonly exists in VRD settings. By re-balancing the proposals, SABRA significantly improves the overall performance of VRD algorithms.

\begin{figure*}[t]
\centering
\includegraphics[width=.9\linewidth]{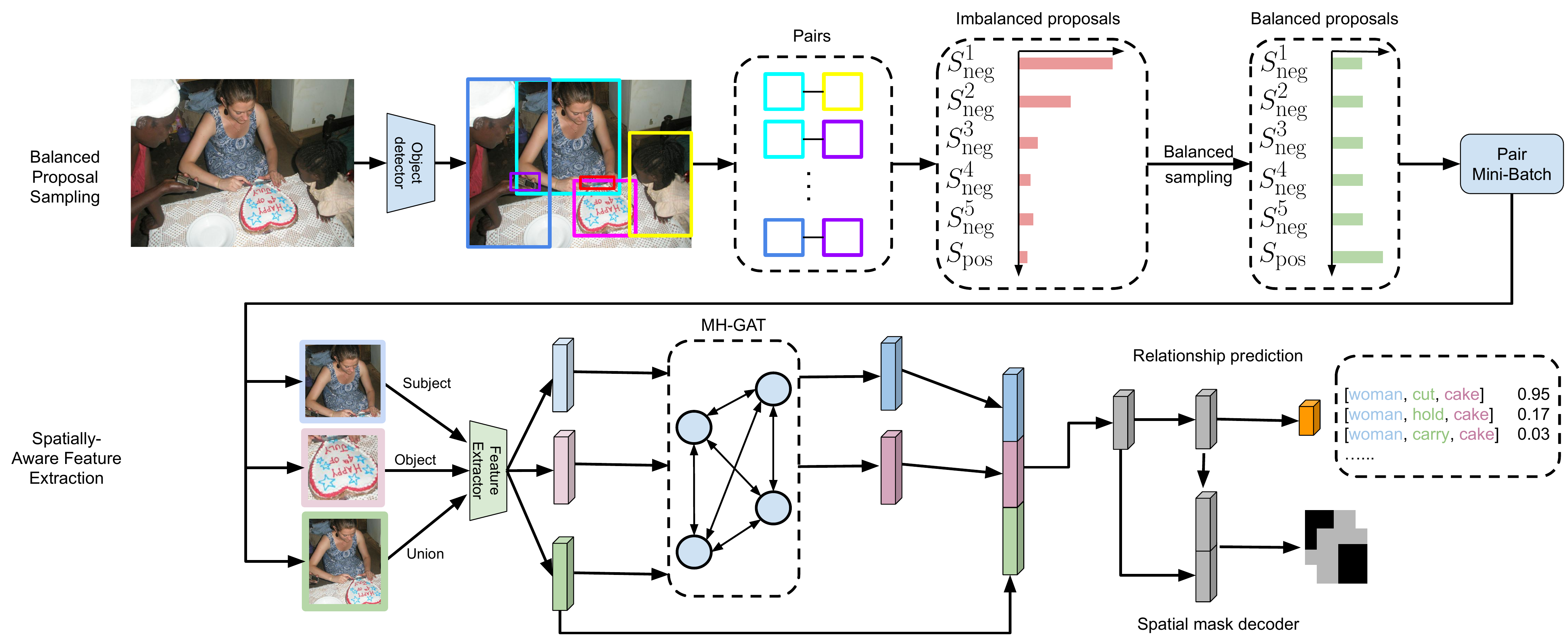}
\caption{SABRA classifies the false positive [subject, object] proposals into 5 sub-categories $\sneg^{1:5}$ and use Balanced-Sampling to re-balance the categorical distribution $\textrm{Cat}(\spos, \sneg^{1:5})$. To reduce the false positive relationship predictions caused by spatial ambiguity, SABRA first uses MH-GAT to capture the global context and then learns to predict the mask of [subject, object].}
\label{fig:sabra}
\end{figure*}

\subsection{Spatial Information}
Spatial information is key to understanding the relationship between objects, e.g., ``woman sitting on the chair". 
Prior methods fuse spatial information with positional embedding, which normalizes the absolute or relative coordinates of the subject, object, and union bounding boxes~\cite{DBLP:conf/cvpr/ZhangSETC19}. 
However, simple positional embedding implicitly captures spatial information with position coordinates as inputs to networks, which is unable to capture the explicit spatial configuration in the feature space. \cite{DBLP:conf/icip/GkanatsiosPKZM19} introduces binary masks which explicitly specify the subject and object positions, 
implicitly specifying the spatial configuration by concatenating with union features.
With the recent advances in graph neural networks, the relevant positional information can be captured by message passing between instances~\cite{DBLP:conf/mm/ZhouZH19}. The implicit message passing, nevertheless, loses the contextual grounding of the \subobj pair.
In contrast, the spatial mask decoder learns to predict the positional information of the \subobj, while capturing the relevant features by end-to-end learning.

\section{SABRA}
We introduce Spatially-Aware Balanced negative pRoposal sAmpling (SABRA) framework for effective visual relationship detection against false positives. The SABRA framework is shown in Fig.~\ref{fig:sabra}. SABRA simplifies the difficult optimization landscape of VRD tasks caused by imbalanced proposal distribution by Balanced Negative Proposal Sampling (BNPS) scheme. BNPS divides the imbalanced negative proposals into sub-classes and creates balanced batches for model optimization. In addition, SABRA further improves the existing GNN-based spatial feature extraction pipeline from two perspectives: (1) a heterogeneous message passing scheme for global scene understanding; (2) a spatial mask decoder for local spatial configuration learning. SABRA combines these objectives and achieves the SOTA performance on multiple datasets.

\subsection{Balanced Negative Proposal Sampling}
\subsubsection{Negative Relationship Proposals}\label{sect:neg_rel_prop}
Despite abundant researches in improving feature extraction for positive \subobjrel pairs in VRD, effective learning under a large number of negative proposals is never explored, which has a greater impact on the performance of the VRD task. 

Suppose we acquire two sets of bounding boxes $B_{\textrm{subject}},B_{\textrm{object}}$ from the object detector, where $B_\textrm{subject}$ contains all the bounding boxes for subjects, and $B_{\textrm{object}}$ contains all the bounding boxes which probably contain an object. 
A relationship proposal is defined as a 
tuple $(b_1, b_2)$. We define the set of relationship proposals $S$ as follows:

\begin{equation}
    S=\{(b_1, b_2) \mid b_1 \in B_{\textrm{subject}},b_2 \in B_{\textrm{object}}\}.
\end{equation}

$S$ can be divided into two disjoint subsets $\spos$ and $\sneg$,

\begin{equation}
    S = \spos\cup \sneg,\quad \spos \cap \sneg = \emptyset,
\end{equation}
where $\spos$ denotes the set of positive proposals that correspond to the ground truth, and $\sneg$ stands for the set of wrong proposals, i.e., negative proposals. Specifically,
\begin{align}
    \spos = \{(b_1,& b_2)\mid \exists (g_1, g_2)\in \textrm{GT}_\textrm{asso},\nonumber\\ 
    &\min (\textrm{IoU}(b_1, g_1), \textrm{IoU}(b_2, g_2)) \geq 0.5\},
\end{align}
where $\textrm{GT}_\textrm{asso}$ is the set of ground truth \subobj pairs and IoU stands for the intersection of the union.

Given that common VRD algorithms choose top-$K$ detections with highest confidence to generate relationship proposals, we have $|S| = K^2$. However, consider that there are normally less than 10 positive proposals for an image, i.e., $|\spos| < 10$, when $K=100$, $|\spos| / |S| < 10^{-3}$. Learning under such an extremely imbalanced data distribution is naturally difficult.

One commonly used VRD mini-batch sampling strategy~\cite{DBLP:conf/cvpr/LinDZT20}, inspired by object detection, is to sample 25\% of the mini-batch from $\spos$ and construct the rest 75\% by $\sneg$. Although it ensures the presence of positive proposals, it overlooks the complex distribution of negative proposals $\sneg$.  Different from object detection, negative proposals in VRD are caused mainly by two reasons: inaccurate detections and incorrect associations. 
Inaccurate detections cause negative proposals by generating inaccurate bounding boxes with $\textrm{IoU}(b_1, g_1) < 0.5$. This type of negative proposals can be easily identified by the visual appearance feature of single object detections alone. On the other hand, incorrect associations cause negative proposals in a more complicated way. Due to the limited number of accurate detections $D_\textrm{pos}$, proposals which consist of both correct detections from $D_\textrm{pos}$ only contribute a small portion of $S$. For detections with at least one positive relationship, $D_\textrm{rel} = \{b\mid \exists b', (b, b')\in \spos \vee (b', b) \in \spos \}$, we have $|D_\textrm{rel}| / |D_\textrm{pos}| < 0.2$. Moreover, consider a proposal $(b_1, b_2)$ where $b_1\in D_\textrm{rel}$, i.e., $\exists b'_2, (b_1, b'_2)\in \spos$. The relationship prediction of $(b_1, b_2)$ will be influenced by the positive proposal $(b_1, b'_2)$, which introduces extra confusion.
Thus, the proposals with bounding boxes from $D_\textrm{rel}$ are not only under-represented but more confusing.
In summary, an imbalanced proposal distribution, where the difficulty of each proposal is negatively correlated with its population size, commonly exists in VRD tasks and would degrade the overall performance of VRD algorithms.
\subsubsection{Balanced Negative Proposal Sampling}

Our solution is motivated by learning with data imbalance~\cite{barandela2003restricted,buda2018systematic}. We propose a balanced negative proposal sampling strategy considering 1 positive class and 5 negative classes. We first define two helper functions, $\fbbox$ and $\frel$:

\begin{align}
    &\fbbox(b)= \begin{cases}
		1,&\textrm{if} \;\exists g \in \textrm{GT}_{\textrm{box}}, \textrm{IoU}(b, g) \geq 0.5 , \\
		0,&\textrm{otherwise}.
		\end{cases}\\
	&\frel(b) = \begin{cases}
		1,&\textrm{if} \;\exists (g_1,g_2) \in \textrm{GT}_{\textrm{rel}}, \textrm{maxIoU} (b, (g_1, g_2))  \geq 0.5 , \\
		0,&\textrm{otherwise},\end{cases}
\end{align}
where $\textrm{GT}_\textrm{box}$ denotes the set of ground truth bounding boxes, $\textrm{GT}_\textrm{rel}$ denotes the set of ground truth relationships, and $\textrm{maxIoU}(b, (g_1, g_2)) = \max(\textrm{IoU}(b, g_1), \textrm{IoU}(b, g_2))$. Intuitively, $\fbbox(b)$ indicates if a bounding box $b$ is a positive bounding box, and $\frel(b)$ denotes if bounding box $b$ belongs to a positive relationship. Next, we divide $\sneg$ following these two principles: (1) simple proposals are caused by inaccurate bounding boxes; (2) difficult proposals are introduced by incorrect \subobj associations. We formulate the 5 sub-classes:
\begin{align}
    \sneg^1 =& \{(b_1, b_2) | \neg \fbbox(b_1) \land \neg \fbbox(b_2)\},\nonumber\\
    \sneg^2 =& \{(b_1, b_2) | (\neg \fbbox(b_1) \land \fbbox(b_2))\nonumber\\&\vee (\fbbox(b_1) \land \neg \fbbox(b_2))\},\nonumber\\
    \sneg^3 =& \{(b_1, b_2) | \fbbox(b_1) \land \neg \frel(b_1) \land \fbbox(b_2) \land \neg \frel(b_2)\},\nonumber\\
    \sneg^4 =& \{(b_1, b_2) | (\frel(b_1) \land \fbbox(b_2) \land \neg \frel(b_2)) \vee\nonumber\\ 
    &(\fbbox(b_1) \land \neg \frel(b_1) \land \frel(b_2))\},\nonumber\\
    \sneg^5 =&\{(b_1, b_2) | \frel(b_1) \land \frel(b_2) \land (b_1, b_2)\notin \spos\}.\label{eqn:bnps}
\end{align}
This divides the negative proposals into 5 sub-classes: (1) both detections are incorrect; (2) one detection is incorrect; (3) both detections are correct, but they belong to no \subobjrel triplet; (4) both detections are correct, but only one of them appears in \subobjrel triplets; (5) both detections are correct, but they appear in two disjoint sets of \subobjrel triplets, which is the minority of the negative proposals but is the most difficult to learn.
Together with the $\spos$, we divide $S$ into 6 classes, $S = \bigcup\limits_{i=1}^5\sneg^i \cup \spos$, which better describe the underlying distribution. 

We adopt the most standard \textit{Balanced Sampling} scheme~\cite{DBLP:conf/eccv/ShenLH16} in learning with imbalance literature to address the imbalance of negative proposals $\sneg$, and we introduce Balanced Negative Proposal Sampling (BNPS). During training time, for each image, we find the top-$K$ object detections, construct $K\times K$ relationship proposals, and count the number of samples for each class. For each proposal $s_i$, we assign a weight $w_i$:
\begin{equation}
    w_i = \begin{cases}
    0.25 / |\spos|,&\textrm{if} \; s_i\in \spos,\\
    0.15 / |\sneg^{j}|,&\textrm{if} \; s_i \in \sneg^j , j=1, \dots, 5
    \end{cases}
\end{equation}
where we keep the weight, 0.25, of positive proposals, while re-balancing the weight of negative proposals, which improves the prediction of low-frequency difficult classes.

\subsection{Spatially-Aware Embedding Learning}
For low-frequency difficult false positives, GNNs generally improve the global scene understanding~\cite{xu2019learning,mi2020hierarchical,li2019transferable} and learning the spatial position of objects improves the local feature representation of \subobj pairs~\cite{DBLP:conf/cvpr/ZhangSETC19,DBLP:conf/cvpr/LiangLX17,DBLP:conf/mm/ZhouZH19}. For global features, SABRA improves the graph representation learning in VRD by Multi-head Heterogeneous Graph Attention networks (MH-GAT) for message passing. For local features, SABRA uses a Spatial Mask Decoder (SMD) to explicitly model the positional configuration of a \subobj pair. 

\subsubsection{Multi-head Heterogeneous Graph Attention}
We introduce Multi-head Heterogeneous Graph Attention Networks (MH-GAT) for effective learning of the global context. One limitation of widely used Graph Attention networks is that it assumes a homogeneous node feature distribution and uses the same set of parameters for learning embeddings for all nodes~\cite{velickovic2018graph}. However, the features of subjects, objects, and union bounding boxes often follow different distributions. Directly using GAT might degrade the final performance. MH-GAT extends GAT by performing message passing on heterogeneous nodes, i.e., subject, object, and union feature, in VRD. We treat each object in the image as a node $i$ with feature $x_i$, and construct a fully connected graph $G=(V, E)$, where $|V|=N$ is the set of nodes and $E=\{(i,j)\mid i,j = 1,2,\dots,N\}$. We treat the feature for edge $(i,j)$ as the union feature $e_{ij}$. The MH-GAT message passing mechanism is as follows:
\begin{align}
    x'_i &= x_i + \sum\limits_{j\in N(i)\cup\{i\}} \alpha_{ij} f_m ([x_i, x_j, e_{ij}]),\\
    \alpha_{ij} &= \frac{\exp (g_{ij})}{\sum\limits_{k\in N(i)\cup \{i\}}\exp (g_{ik})},\\
    g_{ij} &= [f_s(x_i), f_t(x_j), f_e(e_{ij})],
\end{align}
where $\alpha_{ij}$ is the attention weight for edge $(i,j)$, $f_m$, $f_s$, $f_t$ and $f_e$ are functions for message embedding, source node feature embedding, target node feature embedding, and edge feature embedding respectively. 
In contrast to most of the GNNs, when computing the attention values, we process source features, target features, edge features with different functions. This allows modeling the heterogeneous distribution between different nodes.
This formulation also enables our model to pass diverse features from a single source node to different target nodes. Besides, we apply a multi-head attention mechanism, which provides a better modeling ability to capture the complex attentions among objects. 
\begin{table*}[t]
    \centering
\fontsize{8}{10}\selectfont
\begin{tabular}{ccccccccc}
\toprule
\multicolumn{2}{c}{} & V-COCO & \multicolumn{6}{c}{HICO-DET}\\  &  &  & \multicolumn{3}{c}{Default}  & \multicolumn{3}{c}{Known Objects}    \\
\cmidrule(lr){4-6}\cmidrule(lr){7-9}
Method   & Backbone & AP$_{role}$   & Full  & Rare  & Non-Rare  & Full  & Rare  & Non-Rare      \\
\midrule
\midrule
iCAN~\cite{DBLP:conf/bmvc/GaoZH18}  & ResNet50      & 45.30      & 14.84         & 10.45     & 16.15         & 16.26         & 11.33     & 17.73     \\
Contextual Attention~\cite{DBLP:conf/iccv/WangAKKP0L19} & ResNet50      & 47.30      & 16.24         & 11.16     & 17.75         & 17.73         & 12.78     & 19.21     \\
In-GraphNet~\cite{DBLP:conf/ijcai/YangZ20}     & ResNet50       & 48.90      & 17.72     & 12.93    & 13.91     & -     & -     & -     \\
VCL~\cite{DBLP:journals/corr/abs-2007-12407} & ResNet50       & 48.30      & 19.43     & \textbf{16.55}    & 20.29     & 22.00     & 19.09    & 22.87    \\
PD-Net~\cite{DBLP:journals/corr/abs-2008-02918} & ResNet50 & 52.30 & 20.76 & 15.68 & 22.28 & 25.58 & 19.93 & 27.28 \\
SABRA(Ours)     & ResNet50             & \textbf{53.57} & \textbf{23.48}       & 16.39                 & \textbf{25.59}       & \textbf{28.79}       & \textbf{22.75}       & \textbf{30.54}       \\
\midrule
InteractNet~\cite{DBLP:conf/cvpr/GkioxariGDH18} & ResNet50-FPN & 40.00 & 9.94 & 7.16 & 10.77 & - & - & - \\
PMFNet~\cite{DBLP:conf/iccv/WanZLLH19} & ResNet50-FPN         & 52.00             & 17.46      & 15.65    & 18.00    & 20.34     & 17.47     & 21.20      \\
DRG~\cite{DBLP:conf/eccv/GaoXZH20} & ResNet50-FPN         & 51.00             & 19.26      & \textbf{17.74}    & 19.71    & 23.40     & 21.75     & 23.89      \\
IP-Net~\cite{wang2020learning} & ResNet50-FPN         & 51.00             & 19.56      & 12.79    & 21.58    & 22.05     & 15.77    & 23.92      \\
Contextual HGNN~\cite{DBLP:journals/corr/abs-2010-10001} & ResNet50-FPN         & 52.70             & 17.57      & 16.85    & 17.78    & 21.00     & 20.74    & 21.08      \\
SABRA(Ours)  & ResNet50-FPN         & \textbf{54.69} & \textbf{24.12} & 15.91 & \textbf{26.57} & \textbf{29.65} & \textbf{22.92} & \textbf{31.65} \\
\midrule
VSGNet~\cite{DBLP:conf/cvpr/UlutanIM20}     & ResNet152  & 51.76   & 19.80  & 16.05   & 20.91  & -   & -    & -     \\
PD-Net~\cite{DBLP:journals/corr/abs-2008-02918} & ResNet152 & 52.20 & 22.37 & \textbf{17.61} & 23.79 & 26.89 & 21.70 & 28.44 \\
ACP~\cite{DBLP:journals/corr/abs-2007-08728}    & ResNet152   & 52.98    & 20.59   & 15.92    & 21.98  & -  & -   & -    \\
SABRA(Ours)  & ResNet152            & \textbf{56.62} & \textbf{26.09} & 16.29 & \textbf{29.02} & \textbf{31.08} & \textbf{23.44} & \textbf{33.37}\\
\bottomrule
\end{tabular}
    \caption{Results on V-COCO and HICO-DET datasets. In most of the cases, SABRA significantly outperforms SOTA methods.}
	\label{tab:hoi}
\end{table*}
\begin{table*}[t]
	\centering
\fontsize{8}{10}\selectfont
	\begin{tabular}{cccccccccc}
	    \toprule
        \multicolumn{2}{c}{~} & \multicolumn{4}{c}{Relationship Detection} & \multicolumn{4}{c}{Phrase Detection} \\
        \multicolumn{2}{c}{~} & \multicolumn{2}{c}{k=1} & \multicolumn{2}{c}{k=70} & \multicolumn{2}{c}{k=1} & \multicolumn{2}{c}{k=70} \\
        \cmidrule(lr){3-4}\cmidrule(lr){5-6}\cmidrule(lr){7-8}\cmidrule(lr){9-10}
        Method & Backbone & R@50 & R@100 & R@50 & R@100 & R@50 & R@100 & R@50 & R@100 \\
        \hline
        \hline
        \multicolumn{1}{c}{VRD~\cite{DBLP:conf/eccv/LuKBL16}} & VGG16   & 17.03 & 16.17 & 24.90 & 20.04 & \multicolumn{1}{c}{14.70} & 13.86 & 21.51 & 17.35 \\
        \multicolumn{1}{c}{KL distilation~\cite{DBLP:conf/iccv/YuLMD17}} & VGG16   & 19.17 & 21.34 & 22.68 & 31.89 & \multicolumn{1}{c}{23.14} & 24.03 & 26.32 & 29.43 \\
        \multicolumn{1}{c}{Zoom-Net~\cite{DBLP:conf/eccv/YinSLYWSL18}} & VGG16   & 18.92 & 21.41 & 21.37 & 27.30 & \multicolumn{1}{c}{24.82} & 28.09 & 29.05 & 37.34 \\
        \multicolumn{1}{c}{CAI + SCA-M~\cite{DBLP:conf/eccv/YinSLYWSL18}} & VGG16   & 19.54 & 22.39 & 22.34 & 28.52 & \multicolumn{1}{c}{25.21} & 28.89 & 29.64 & 38.39 \\
        \multicolumn{1}{c}{Hose-Net~\cite{DBLP:journals/corr/abs-2008-05156}} & VGG16   & 20.46 & 23.57 & 22.13 & 27.36 & \multicolumn{1}{c}{27.04} & 31.71 & 28.89 & 36.16 \\
        \multicolumn{1}{c}{RelDN~\cite{DBLP:conf/cvpr/ZhangSETC19}} &  VGG16   & 18.92 & 22.96 & 21.52 & 26.38 & \multicolumn{1}{c}{26.37} & 31.42 & 28.24 & 35.44 \\
        \multicolumn{1}{c}{AVR~\cite{lv2020avr}} & VGG16   & 22.83 & 25.41 & 27.35 & 32.96 & \multicolumn{1}{c}{29.33} & 33.27 & \textbf{34.51} & 41.36 \\
        \multicolumn{1}{c}{SABRA(Ours)} & VGG16   & \textbf{24.47} & \textbf{29.16} & \textbf{27.27} & \textbf{33.99} & \textbf{30.57} & \textbf{36.80} & 33.39 & \textbf{41.79} \\
        \midrule
        \multicolumn{1}{c}{GPS-Net~\cite{DBLP:conf/cvpr/LinDZT20}} & VGG16 (MS COCO) & 21.50 & 24.30 & - & - & \multicolumn{1}{c}{28.90} & 34.00 & - &-\\
        \multicolumn{1}{c}{MCN~\cite{zhan2020multi}} & VGG16 (MS COCO) & 24.50 & 28.00 & - & - & \multicolumn{1}{c}{31.80} & 37.10 & - & - \\
        \multicolumn{1}{c}{SABRA(Ours)} & VGG16 (MS COCO) & \textbf{26.29} & \textbf{31.08} & \textbf{29.44} & \textbf{36.44} & \textbf{32.01} & \textbf{38.48} & \textbf{35.45} & \textbf{44.07} \\
        \midrule
        \multicolumn{1}{c}{UVTransE~\cite{hung2020contextual}} & VGG16 (VG) & 25.66 & 29.71 & 27.32 & 34.11 & \multicolumn{1}{c}{30.01} & 36.18 & 31.51 & 39.79 \\
        \multicolumn{1}{c}{SABRA(Ours)} & VGG16 (VG) & \textbf{27.87} & \textbf{32.48} & \textbf{30.71} & \textbf{37.71} & \textbf{33.56} & \textbf{39.62} & \textbf{36.62} & \textbf{45.29} \\
        \midrule
        \multicolumn{1}{c}{ATR-Net~\cite{gkanatsios2019attention}} & ResNet101 & - & - & - & - & \multicolumn{1}{c}{31.96} & 36.54 & 36.06 & 43.45 \\
        
        \multicolumn{1}{c}{SABRA(Ours)} & ResNet101 & \textbf{26.73} & \textbf{31.11} & \textbf{29.92}  & \textbf{37.43}  & \textbf{32.81}  & \textbf{38.68}  & \textbf{36.24}  & \textbf{45.26} \\

        \bottomrule
        
	\end{tabular}
	\caption{Results on the VRD dataset. SABRA outperforms all SOTA methods by a large margin in all cases. By default, all backbones are pretrained with ImageNet. To align with the setups of some prior works~\cite{DBLP:conf/cvpr/LinDZT20,zhan2020multi,hung2020contextual}, we use MS COCO or Visual Genome as additional data for training, denoted by VGG (MS COCO) or VGG (VG).}
	\label{tab:vrd}
\end{table*}
\subsubsection{Spatial Mask Decoder}
\begin{figure}[!htb]
\centering
\includegraphics[width=0.8\linewidth]{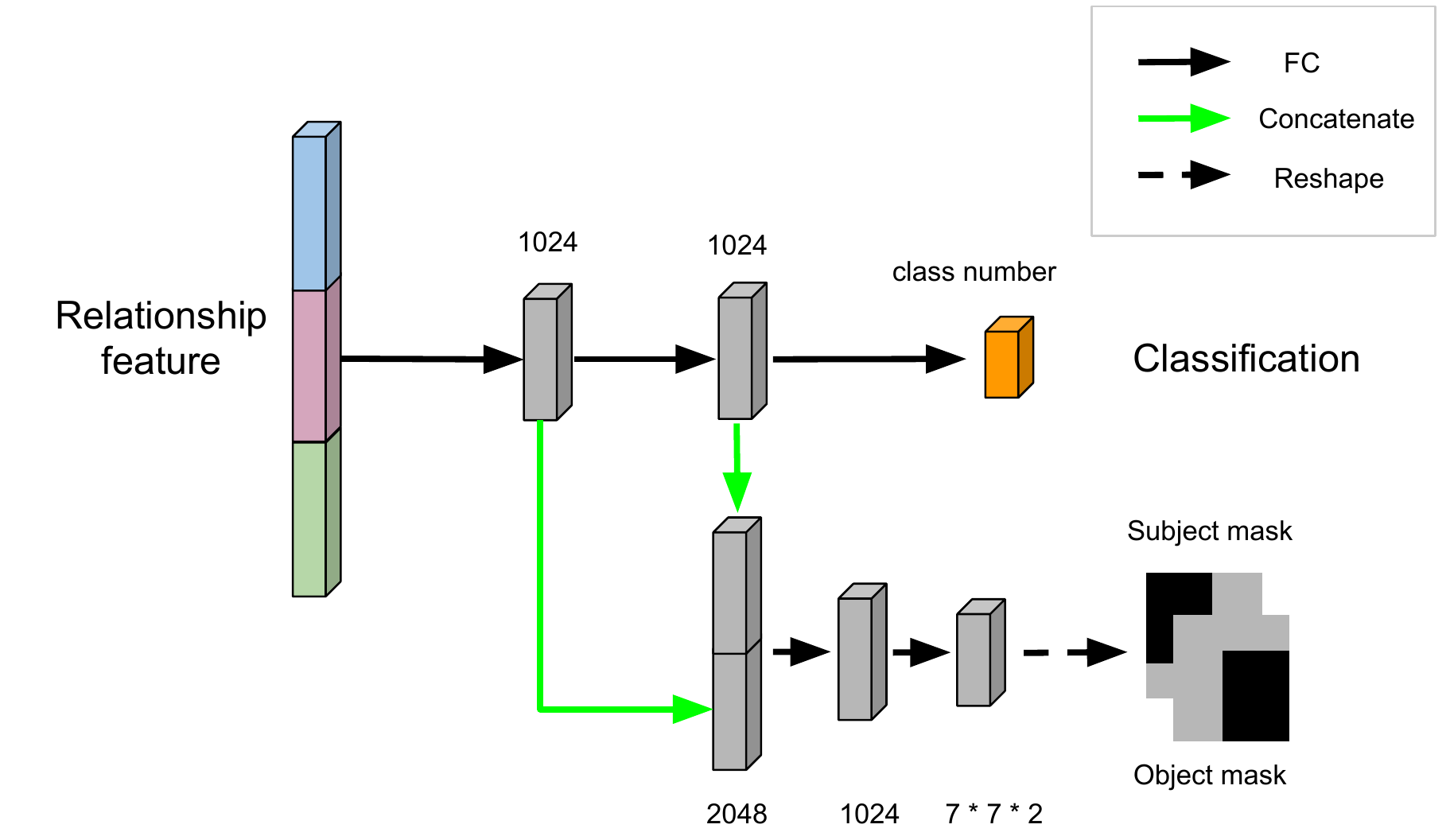}
\caption{Structure of Spaital Mask Decoder}
\label{fig:spatial}
\end{figure}

To learn the local spatial configuration, we present Spatial Mask Decoder (SMD) which predicts the locations of \subobj. The architecture of SMD is given in Fig.~\ref{fig:spatial}. SMD predicts a $2\times l_p\times l_p$ mask using the VRD feature vectors, where $l_p$ represents the pooling size and the two channels represent the spatial location of subjects and objects in the pooled feature map with size $l_p$. This guarantees that the local spatial configuration is tightly embedded in the feature vector for relationship prediction. We scale the absolute coordinates of the \subobj pair to get the relative position in the pooled feature map. Compared to reconstruction in the image space, the ROI feature space preserves the locality of the union bounding box but requires fewer parameters. Different from the standard positional embedding~\cite{DBLP:conf/cvpr/ZhangSETC19} which implicitly utilizes the spatial information with the absolute positions as an input, SMD explicitly learns a structured embedding. Compared to ~\cite{DBLP:conf/icip/GkanatsiosPKZM19} which concatenates binary mask as position feature, SMD explicitly imposes the spatial structure in the embedding vector and gives better spatially-aware embeddings.
We empirically show SMD outperforms these variants in our experiments.

\subsection{SABRA Framework}
The pipeline of SABRA is shown in Fig.~\ref{fig:sabra}. The input image is fed into an object detector to predict all bounding boxes, which are exhaustively paired to generate all relationship proposals.  
In the proposal sampling stage, given the imbalanced relationship proposals, SABRA constructs a balanced pair mini-batch by BNPS that samples data point $i$ proportionally to its weight $w_i$. The features of subjects, objects, and union bounding boxes are fed into MH-GAT with heterogeneous message passing to extract spatial object interactions. We use only one layer of MH-GAT to avoid over-smoothing in GNNs~\cite{yu2020spatio}. The mask decoder discovers the spatial configuration of \subobj pairs in the learned embeddings. SABRA is robust to the imbalanced negative proposal distribution and reduces the number of false positive predictions caused by spatial ambiguity. More details are available in the appendix.

During training, we jointly optimize the object detector and relationship classifier by
\begin{equation}
L = L_{\textrm{RPN}}^{\textrm{cls}} + L_{\textrm{RPN}}^{\textrm{loc}} + L_{\textrm{RCNN}}^{\textrm{cls}} + L_{\textrm{RCNN}}^{\textrm{loc}}+L_{\textrm{VRD}}^{\textrm{cls}} + L_{\textrm{VRD}}^{\textrm{mask}},
\end{equation}
where $L_{\textrm{RPN}}^{\textrm{cls}}$, $L_{\textrm{RPN}}^{\textrm{loc}}$, $L_{\textrm{RCNN}}^{\textrm{cls}}$, $L_{\textrm{RCNN}}^{\textrm{loc}}$ are losses for object detection. Their definitions are the same as ~\cite{ren2015faster}.$L_{\textrm{VRD}}^{\textrm{cls}}$ is a classification loss of each relation proposal. $L_{\textrm{VRD}}^{\textrm{mask}}$ is an auxiliary loss used in our SMD model. These two losses are designed for the VRD branch. We opt for Binary Cross Entropy loss for both items.
During inference, SABRA predicts all given relationship proposals without sampling.

\begin{table}[!htb]
\centering
\fontsize{8}{10}\selectfont
\begin{tabular}{ccccc}
\toprule
 & Sampling  & Spatial Learning & GNN            & AP$_{role}$    \\
\hline
\hline

1                    & -         & -                & -              & 50.20           \\
2                    & BNPS      & -                & -              & 52.24          \\
3                    & -         & SMD              & -              & 50.84          \\
4                    & -         & -                & MH-GAT         & 51.65          \\
\midrule
5                    & -         & SMD              & MH-GAT         & 52.82          \\
6                    & BNPS-2cls & SMD              & MH-GAT         & 53.74          \\
7                    & BNPS-3cls & SMD              & MH-GAT         & 53.93          \\
\midrule
8                    & BNPS      & -                & MH-GAT         & 53.67          \\
9                    & BNPS      & Binary~\cite{DBLP:conf/icip/GkanatsiosPKZM19}           & MH-GAT         & 53.90           \\
10                   & BNPS      & PE~\cite{DBLP:conf/cvpr/ZhangSETC19}               & MH-GAT         & 54.29          \\
\midrule
11                   & BNPS      & SMD              & -              & 52.53          \\
12                   & BNPS      & SMD              & M-GAT          & 51.65          \\
13                   & BNPS      & SMD              & MH-GAT-no-edge & 53.80           \\
\midrule
14                   & BNPS      & SMD              & MH-GAT         & \textbf{54.69}\\
\bottomrule
\end{tabular}
\caption{Ablation Study on V-COCO dataset. We compare each model variant with the baseline (row 1) and the SABRA (row 14).
}
\label{tab:ablation}
\end{table}
\section{Experiment}
We evaluate SABRA on three commonly used datasets: V-COCO~\cite{DBLP:journals/corr/GuptaM15}, HICO-DET~\cite{DBLP:conf/wacv/ChaoLLZD18}, and VRD~\cite{DBLP:conf/eccv/LuKBL16}, covering both human-object interaction (V-COCO and HICO-DET) and general object relationship detection (VRD). We compared SABRA over 20 SOTA methods. Specifically, we trained SABRA with different backbones and perform a comprehensive and fair comparison with existing methods. 
We incrementally add our proposed components to a baseline model. The baseline directly concatenates appearance features of subjects, objects, and their union features, which are fed into a 3-layer Multi-Layer Perceptron (MLP) for classification.

As a brief conclusion, we show that: 1) SABRA generally outperforms other methods by a large margin; 2) the balanced negative proposal sampling strategy can reduce the number of false positive predictions; 3) spatial mask decoder successfully reduces the number of false positives caused by spatial ambiguity.

\subsection{Datasets}

V-COCO is based on the 80-class object detection annotations of COCO~\cite{DBLP:conf/eccv/LinMBHPRDZ14}. 
It has 10,346 images (2,533 for training, 2,867 for validating and 4,946 for testing). HICO-DET has a total of 47,774 images, covering 600 categories of human-object interactions over 117 common actions on 80 common objects. VRD dataset contains 4,000 images in the train split and 1,000 in the test split. It has 100 different types of objects and 70 types of relationships. 

\subsection{Evaluation Metrics}

We followed the convention in prior literature and used three different evaluation metrics for these datasets.

Following \cite{DBLP:conf/eccv/LuKBL16}, we use Recall@$N$ as our evaluation metric on VRD. Recall@$N$ computes the \emph{recall} rates using the top $N$ predictions per image. To be consistent with prior works, we report Recall@50 and Recall@100 in our experiments. We evaluate two tasks: \textit{relationship detection} that outputs triple labels and evaluates bounding boxes of the subject and object separately; \textit{phrase recognition} that takes a triplet as a union bounding box and predicts the triple labels.
Besides, we report the top-$k$ predictions as ~\cite{DBLP:conf/iccv/YuLMD17}, where $k = 1$ or $k = 70$.

For V-COCO and HICO-DET, mean average precision (mAP) is used to estimate the performance. A triplet [human, verb, object] is considered a positive prediction if and only if there exists a triplet [human', verb', object'] in the ground truth satisfying: (1) IoU(human, human') $\geq 0.5$, (2) verb = verb', (3) IoU(object, object') $\geq 0.5$

In HICO-DET, we calculate the mAP among all pairs of [verb, object]. In \emph{Default} mode, we calculate AP on all images. In \emph{Known Objects} mode, the category of the bounding box is known and we only calculate AP between humans and objects from a specific category. In V-COCO, we calculate the mAP for all categories of verbs, which is called AP$_{role}$. More details are available in the appendix.

\subsection{Quantitative Results}

We present our results in Table~\ref{tab:hoi} for HOI datasets (V-COCO and HICO-DET) and Table~\ref{tab:vrd} for the VRD dataset. For HOI, we cluster our results according to the backbones, including ResNet50, ResNet50-FPN, and ResNet152, with increasing feature extraction ability. For VRD datasets, VGG-16 pretrained on ImageNet is used for all methods. To align with the setup of some prior works~\cite{DBLP:conf/cvpr/LinDZT20,zhan2020multi,hung2020contextual}, we use MS COCO and VG as additional datasets for VGG16.

We observe that SABRA generally improves the SOTA methods significantly on all datasets for both HOI and VRD. For example, on V-COCO with ResNet152 backbone, SABRA achieves 56.62 mAP while the SOTA model, ACP~\cite{kim2020detecting} gives an mAP of 52.98. Specifically, we want to highlight the performance gain of SABRA on the V-COCO dataset. V-COCO uses the object annotations of the COCO dataset, which are accurate and include all objects, regardless of the relationships. This gives a more imbalanced distribution than HICO-DET, where only the objects in relationships are considered. 
Furthermore, accurate annotations also give rise to better object detection, which amplifies the significance of spatial information for a good performance.
Specifically, we noticed that for PD-Net~\cite{DBLP:journals/corr/abs-2008-02918}, using a more powerful backbone (ResNet152) gives no performance gain than the smaller one (ResNet50). SABRA, on the contrary, can fully exploit the feature extraction power of ResNet152 and give a large performance improvement compared with ResNet50 (56.62 V.S. 53.57).
\begin{figure*}[t]
	\centering
	
	\begin{tabular}{c c c}
    \includegraphics[width=0.29\linewidth]{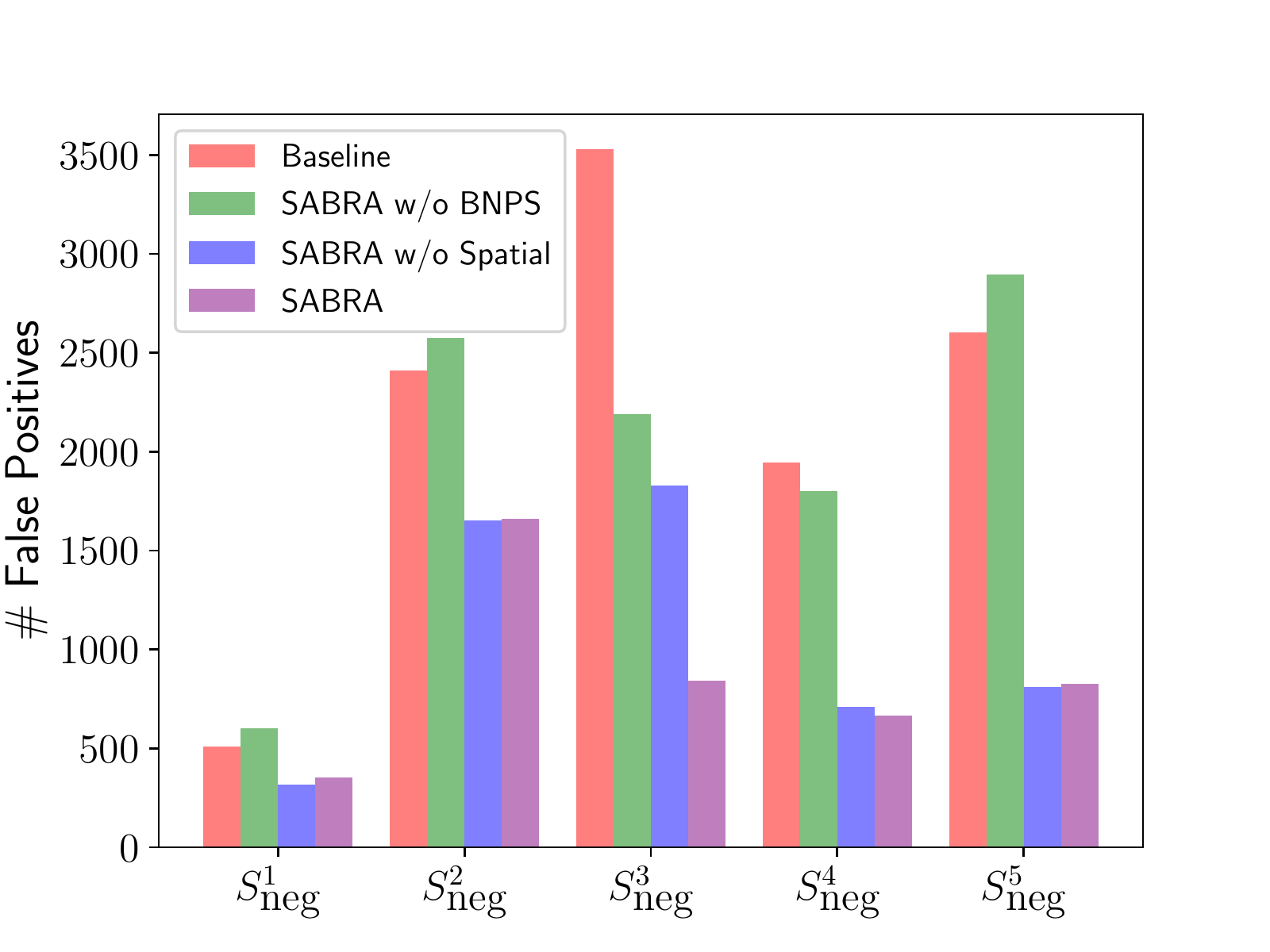} &
    \includegraphics[width=0.25\linewidth]{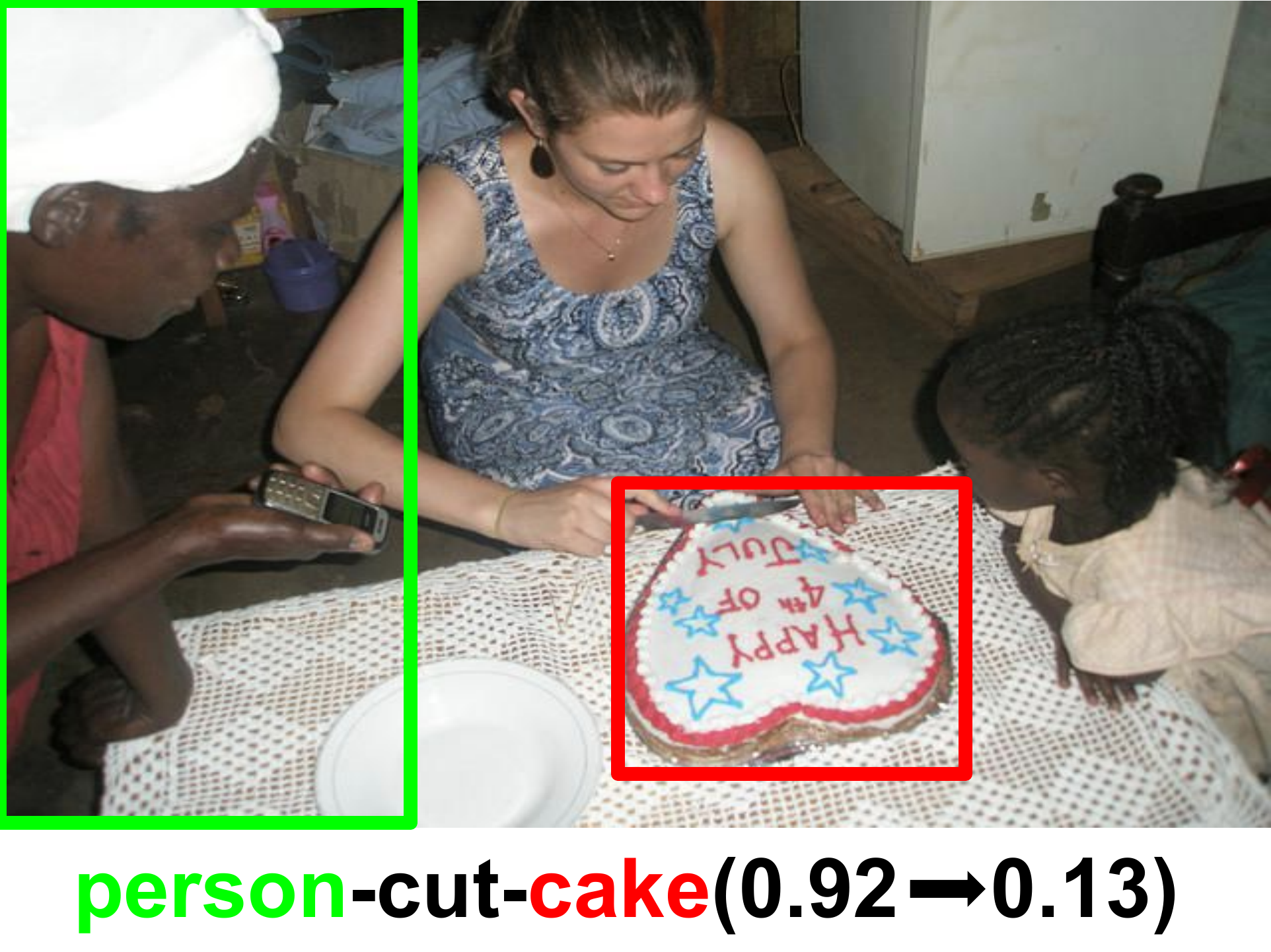} &
    \includegraphics[width=0.25\linewidth]{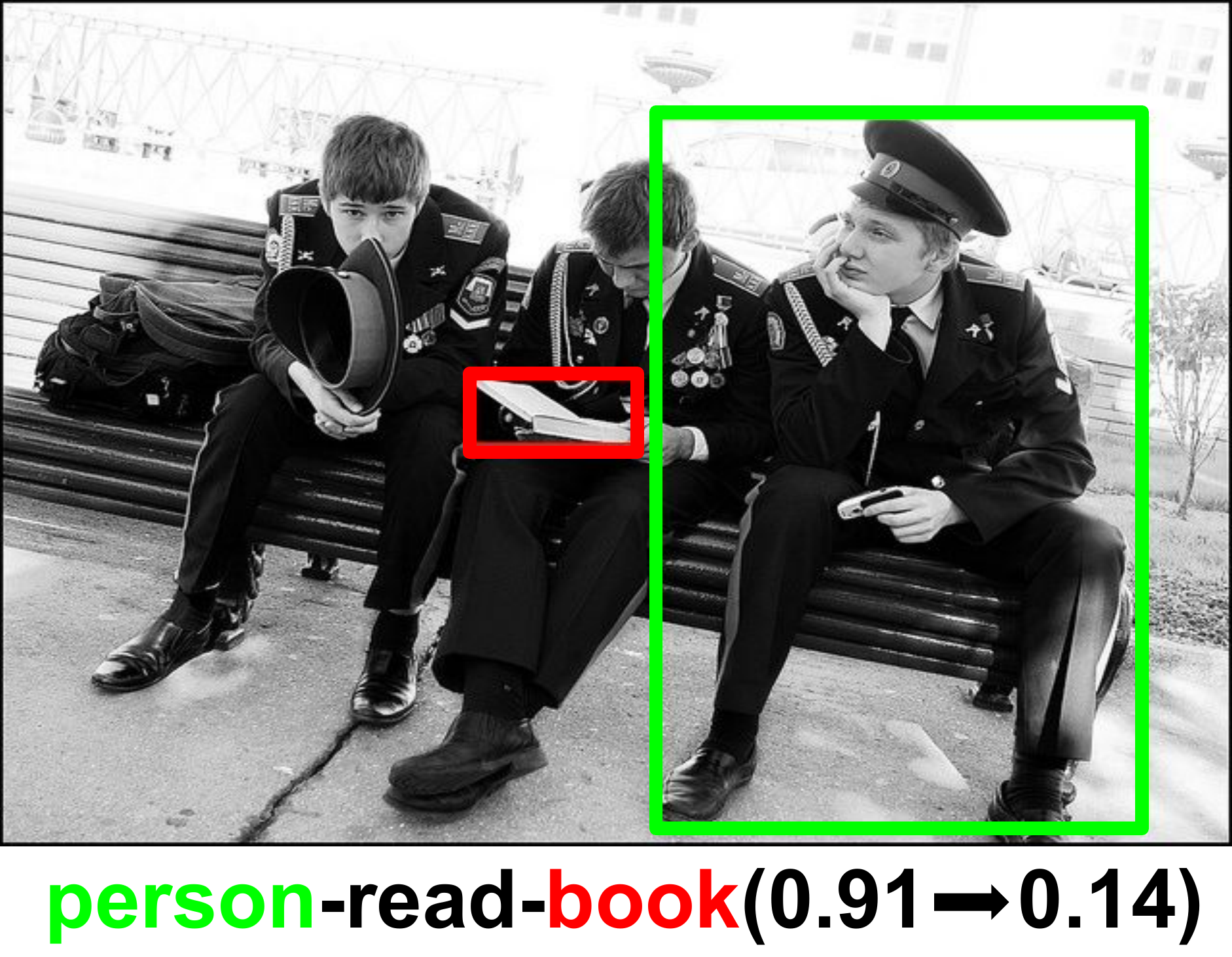}\\
    (a) False positive comparison & \multicolumn{2}{c}{(b) False positive reduction example of SABRA }
	\end{tabular}
	\centering
	\caption{Qualitative analysis of SABRA on V-COCO dataset. (a) The number of false positive predictions under each class by Baseline model, SABRA w/o BNPS, SABRA w/o spatial module (MH-GAT + SMD), and SABRA. BNPS and spatial module significantly reduce the number of false positive predictions.
	(b) Predictions of SABRA. In $\textrm{sub-predicate-obj} (v_1, v_2)$, $v_1$ denotes the prediction confidence of the baseline model and $v_2$ denotes the confidence of SABRA. SABRA successfully reduced the low-frequency difficult false positives.
	}
	\label{fig:vis}
\end{figure*}

\subsection{Ablation Studies}
We conduct a comprehensive ablation study on the V-COCO dataset to understand each proposed component: Balanced Negative Proposal Sampling (BNPS), Multi-head Heterogeneous Graph Attention (MH-GAT), and Spatial Mask Decoder (SMD). We present the results in Table~\ref{tab:ablation}, where row 1 denotes the baseline model and row 14 denotes SABRA with all proposed components.

\textbf{Each proposed component improves the baseline}. We perform incremental analysis in rows 1-4. Specifically, we observe that by simply improving the optimization process with BNPS, baseline + BNPS (row 2) gains 2.04 improvement on AP$_{role}$. This suggests that imbalanced proposal distribution significantly hinders the model performance, and BNPS addresses this issue effectively.

\textbf{Balancing negative proposal distribution generally improves the VRD performance.}
In rows 5-7 and row 14, we compare BNPS with 3 other alternatives. (1) BNPS-3cls (row 7): balanced sampling over $\{\sneg^1, \sneg^2, \sneg^3\cup \sneg^4\cup \sneg^5\}$, i.e., ignoring the difference between negative proposals when detections are correct; (2) BNPS-2cls (row 6): balanced sampling over $\{\sneg^1\cup\sneg^2, \sneg^3\cup \sneg^4\cup \sneg^5\}$, i.e., further ignoring the differences of negative proposals when detections are incorrect; (3) None (row 5): we remove the BNPS. We fix the positive sample rate to be 25\% and balance the rest 75\% samples over the given distributions. Comparing BNPS with BNPS-2cls and BNPS-3cls, we conclude BNPS improves prediction accuracy for both inaccurate detections and incorrect associations.

\textbf{Understanding spatial information is crucial to VRD.} In rows 8-10 and row 14, we compare the proposed Spatial Mask Decoder (SMD) with 3 alternatives. (1) Binary (row 9)~\cite{DBLP:conf/icip/GkanatsiosPKZM19}: a binary mask over the union feature to specify the position of \subobj; (2) PE (row 10)~\cite{DBLP:conf/cvpr/ZhangSETC19}: naive positional embedding; (3) None (row 8): no spatial learning module. We conclude that: (1) adding spatial information learning generally improves the performance;
(2) Implicitly imposing spatial information with PE or binary masks gives worse performance than SMD which enforces an explicit constraint over the feature space.

\textbf{MH-GAT is more effective in heterogeneous VRD graphs} In rows 11-13 and row 14, we compare MH-GAT with other 4 alternatives: (1) Multi-head GAT~\cite{velickovic2018graph} (row 12), standard homogeneous message passing scheme which has the same number of heads and edge features used in MH-GAT, (2) MH-GAT without edge feature (row 13), and (3) None (row 11). We observe that: (1) GNNs generally improve the VRD performance; (2) by separating the parameters for objects, subjects, and union features, the heterogeneous message passing scheme in MH-GAT improves the M-GAT; (3) edge feature gives extra information and is important to MH-GAT.

\subsection{Qualitative Analysis}
Qualitatively, we provide extra statistics and visualizations on the V-COCO dataset to better understand the performance improvement of SABRA in Fig.~\ref{fig:vis}. 

To verify the ability of SABRA on reducing the number of false positives, we compute the per-image false positive predictions for $\sneg^{1:5}$ by thresholding the prediction confidence at 0.5 in Fig.~\ref{fig:vis}. Compared with the baseline, SABRA w/o BNPS, SABRA w/o Spatial, and SABRA have reduced the total number of low-frequency difficult $\sneg^{3:5}$ by 14.7\%, 58.5\%, and 71.1\%. BNPS gave a sharp decrease on $\sneg^{3:5}$, which suggests the necessity of learning from a balanced proposal distribution. The spatial module successfully filtered the irrelevant objects in $\sneg^3$ and significantly reduced the number of false positives in total. Combining both of them, SABRA gives the best performance among all alternatives. 
However, we also noticed that on $\sneg^{4:5}$ and $\sneg^{1:2}$, the spatial module has no clear improvement, because the spatial module potentially overexploited the correct detection of the positive \subobjrel triplet. Auxiliary constraints on the relationship assignment could be considered to address this issue. We leave it for future study.

We visualize two examples of SABRA successfully reducing the low-frequency difficult false positives where both detections are correct in Fig.~\ref{fig:vis}.b. Without the spatial module, the VRD algorithm assigns high confidence (0.92) to the \subobj pair that the woman is cutting the cake. This confidence was reduced to 0.13 by SABRA.

\section{Conclusion}
We present SABRA for alleviating false positives in VRD. We divided the negative \subobj proposals into 5 sub-classes with imbalanced data distribution, 
and addressed the data imbalance by Balanced Negative Proposal Sampling. SABRA incorporates the global contextual information with MH-GAT and local spatial configuration by SMD. SABRA significantly outperforms the SOTA methods on V-COCO, HICO-DET, and VRD datasets.

As the first paper to consider the data imbalance in the negative proposal distribution, SABRA used a relatively simple strategy, balanced sampling. More advanced techniques, e.g., Meta-Learning~\cite{ren2020balanced}, could be considered.

{\small
\bibliographystyle{ieee_fullname}
%\bibliography{ref}

}

\appendix

\section{Implementation Details}
\subsection{Overall Pipeline}
The overall pipeline of SABRA consists of three major components: backbone, detector, and VRD classifier. Backbone is a feature extractor for object detection and relationship detection. Detector predicts all potential bounding boxes and produces a set of detections:

\begin{align}
B = \{(x_1, y_1, x_2, y_2, \textit{cls}, \textit{score})\},
\end{align}
where $(x_1, y_1)$ are the coordinates of the upper left corner, $(x_2, y_2)$ are the coordinates of the bottom right corner, $\textit{cls}$ is the category of this bounding box, and $\textit{score}$ is the confidence of this prediction. VRD classifier uses this set of detections to generate all relationship proposals $S$ and categorize each proposal. In this way, we obtain a set of triplets associated with a \textit{score} from VRD classifier:

\begin{align}
T = \{(b_1, b_2, \textit{cls}, \textit{score})\}.
\end{align}

\subsection{Backbone}

To fully compare with other methods, we use several backbone setups in experiments including VGG-16, ResNet50, ResNet50-FPN, ResNet101, and ResNet152. In particular, when we use ResNet50-FPN for both object detection and relationship detection following \cite{DBLP:conf/cvpr/GkioxariGDH18}, we use \textit{ResNet50-FPN} to denote this setting. Meanwhile, some other works like \cite{DBLP:journals/corr/abs-2008-02918} use ResNet50-FPN for object detection, but use only ResNet50 for relationship detection. Therefore, we build up this setting named \textit{ResNet50} that a ResNet50 module is shared for both object detection and relationship detection, and an extra FPN is used only for object detection. In addition, FPN is used neither in object detection nor relationship detection process when we use VGG-16, ResNet101, and ResNet152 as the backbone.

Specifically, for V-COCO and HICO-DET datasets, we use ImageNet + MS COCO pre-trained ResNet50, ResNet50-FPN, and ResNet152. For VRD dataset, we respectively use VGG-16 pre-trained on ImageNet, ImageNet + MS COCO, and ImageNet + Visual Genome. We also use ImageNet pre-trained ResNet101 for VRD task.

\subsection{VRD Classifier}

The overall structure of the VRD classifier is shown in Fig. 2 in the main text. In this section, we will describe the training strategy and inference procedure of our VRD classifier. 

\textbf{Training Strategy} VRD classifier receives the detection set $B$ and further divides it into two sets $B_{\textrm{subject}}$ and $B_{\textrm{object}}$. In HOI task, $B_{\textrm{subject}}$ contains all human bounding boxes and $B_{\textrm{object}}$ contains all bounding boxes from detector. In general VRD task, both $B_{\textrm{subject}}$ and $B_{\textrm{object}}$ contain all detection results. Under this definition, we generate a proposal set $S$. For each image, we sample 64 relationship proposals from the set $S$. This hyper-parameter holds across all our model configurations, no matter whether we use BNPS. Moreover, we also keep the ratio of positive proposals as 0.25 for all experiments. Sigmoid activation is used to predict both the confidence of each category and the spatial mask. Correspondingly, binary cross-entropy loss is used for both supervisions. 

\textbf{Inference Procedure} For each relationship proposal $(b_1, b_2)$, we predict the score $s_{\textit{cls}}$ of each relationship category $\textit{cls}$. The final score of triplet $(b_1, b_2, \textit{cls})$ is calculated as below:

\begin{align}
\textit{score} = s_1 \times s_2 \times s_{\textit{cls}},
\end{align}
where $s_1$ and $s_2$ are the scores of bounding box $b_1$ and $b_2$ respectively.

\section{Experiment Setup}\label{sec:setup}

\subsection{Data Preprocessing}

For V-COCO, the ground truth detection is obtained from COCO. As some relationships containing invisible objects, we fill the region of the object with the coordinates as the subject. In other words, we generate a ground truth triplet $(b_1, b_1, \textit{cls})$ when the object is invisible.

For HICO-DET, we generate ground truth detection from all bounding boxes of triplets. It should be noted that, in HICO-DET, the annotation of each relationship is independent. Therefore, there may exist multiple bounding boxes for a single person or object. To generate a unique bounding box annotation for each instance, we merge those bounding boxes where the IoU values are no less than 0.5.

For general VRD, we have no extra pre-processing.

\subsection{Training Procedure}

For V-COCO, we first train the object detection on the COCO dataset. Then we freeze the backbone and jointly train the detector and VRD classifier on the corresponding dataset. 

For HICO-DET, we first train the object detection on the COCO dataset and then finetune on the detection results from HICO-DET. After that, we jointly train the Detector and VRD classifier with a frozen backbone.

For general VRD, we use ImageNet pre-trained VGG-16 backbone to initialize our model and then train object detection on the general VRD dataset. Finally, we jointly train the detector and VRD classifier with the frozen backbone.

In each training process, we use 16 GPUs (1080TI) and train 25 epochs with the initial learning rate being 0.00125. The learning rate is decreased in the 17th and 23rd epoch with a 0.1 decay rate. When training solely the object detection, each batch contains four different images. For the joint training of detection and relationship detection, each batch contains two different images. 

\section{Additional Analysis for BNPS}

\subsection{Other reasons for inaccurate detections}

\begin{table}[ht]
 	\centering
 	\setlength{\tabcolsep}{1.6mm}{
 	\begin{tabular}{ccc}
 	    \toprule
     	\multicolumn{1}{c}{Detection Top-$N$} & RS & BNPS \\
        \hline
        \hline
        100 & 51.95 & 54.29 \\
        50 & 52.82& \textbf{54.69} \\
        40 & 52.53 & 54.43 \\
        30 & 53.05 & 54.52 \\
        20 & 52.92 & 53.68 \\
         \bottomrule
 	\end{tabular}}
 	\vspace{3mm}
 	\caption{AP$_{role}$ performance of the different numbers of detections from the detector on the V-COCO test set. We use ResNet50-FPN as our backbone and modify the sampling strategy and selection of the detector's top-$N$ while keeping others unchanged.}
 	\label{tab:topn}
\end{table}

Although inaccurate bounding boxes lead to a large number of easy negative proposals, we cannot simply ignore these detection results. We observe that decreasing the number of top-$N$ detections may have a negative influence on VRD algorithms. Table ~\ref{tab:topn} is a thorough comparison of the top-$N$ detections we keep from the detector and whether we use BNPS or random sampling (RS) against the model performance. The results suggest that decreasing the number of top-$N$ has no evident improvement in the final performance (Top-50 w/o BNPS V.S. Top-30 w/o BNPS, 52.82 V.S. 53.05). The major reason is that we still need many inaccurate bounding boxes in inference for higher recall values. Besides, we find that our proposed Balanced Negative Proposal Sampling has remarkable improvement no matter which top-$N$ we select. These results strongly prove the effectiveness and robustness of our proposed method. 

\subsection{Analysis of the improvement}

The improvement of BNPS comes from two major sources. Firstly, BNPS reduces the easy negative proposals caused by inaccurate detections; secondly, BNPS balances the difficult negative proposals, considering whether the subject and the object are involved in any triplet from the ground truth. These two parts are both significant and essential, which is partially proven in the ablation study. In this section, we add some extra quantitative and qualitative results.

We include an additional experiment to test if the performance improvement of SABRA comes only from the increase in the number of difficult negative proposals $\sneg^{3:5}$. We compare a BNPS-3cls variant, BNPS-3cls-HN with BNPS. In the original BNPS-3cls, we use a sample rate [0.25, 0.25, 0.25] for $\sneg^1$, $\sneg^2$ and $\sneg^{3:5}$. However, in BNPS, each class receives a sample rate of 0.15, which gives a $0.45$ sample rate for the negative classes $\sneg^{3:5}$. We design BNPS-3cls-HN such that it assigns the same sample rate to the hard negative classes, [0.15, 0.15, 0.45] for $\sneg^1$, $\sneg^2$ and $\sneg^{3:5}$. We train both models on V-COCO and report the results briefly here: BNPS-3cls-HN gives 54.20 AP$_{role}$ while SABRA achieves 54.69. This suggests that the improvement from BNPS-3cls to SABRA is not simply because of the increased sample rate of hard negative proposals. The balance among $\sneg^{3:5}$ also plays a critical role in this improvement.

\begin{figure}[t]
\centering
\includegraphics[width=1\linewidth]{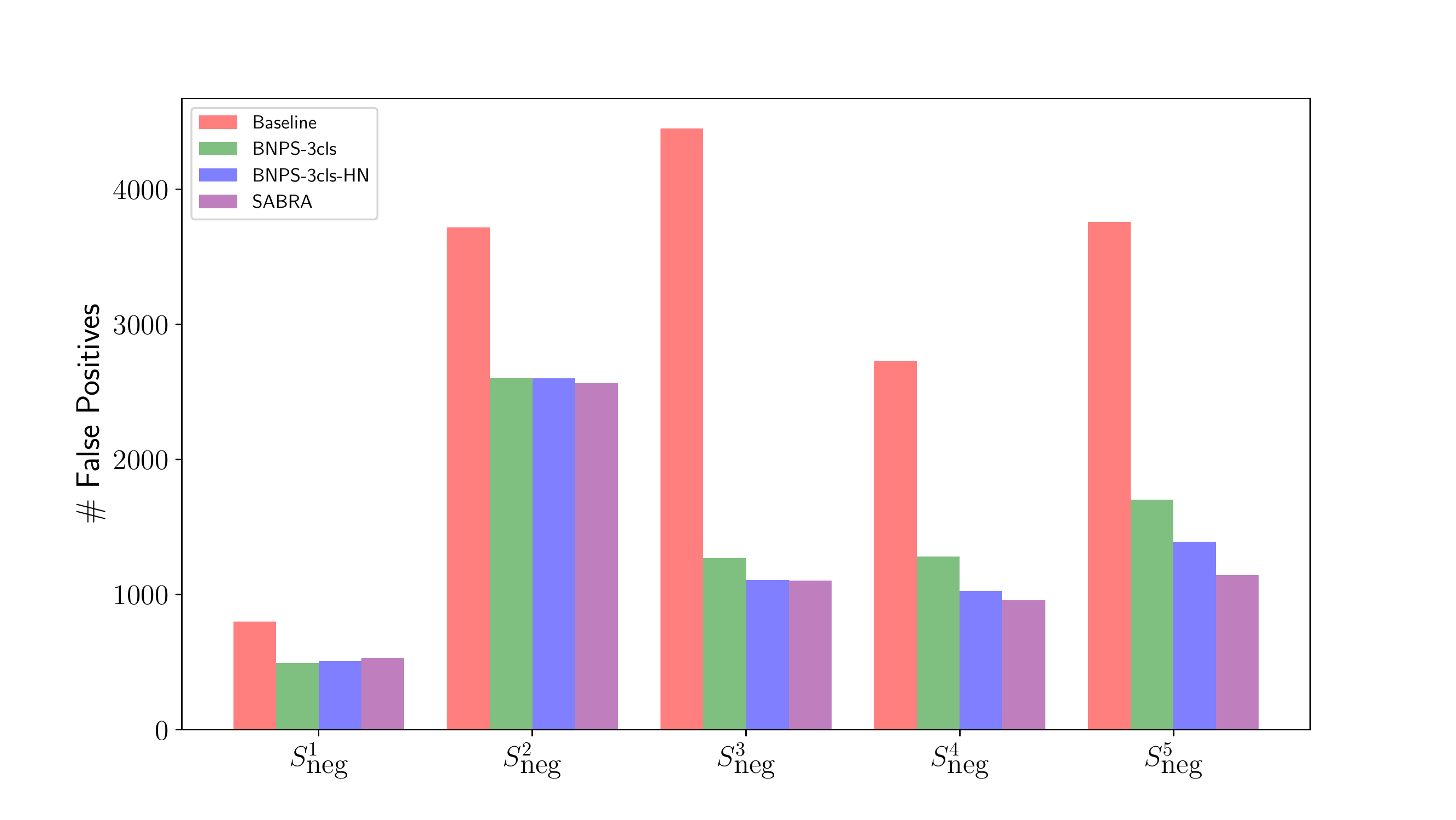}
\caption{Qualitative analysis of different sampling strategies.} 
\label{fig:bnps}
\end{figure}

We also visualize the number of negative predictions of each negative type in Fig.~\ref{fig:bnps}. We observe that: (1) the total number of false positives in low-frequency difficult classes, $\sneg^{3:5}$, of SABRA is lower than BNPS-3cls because we increase the total ratio of difficult negative proposals. (2) the reduction ratio of $\sneg^5$ is higher than that of $\sneg^4$. Meanwhile, the reduction ratio of $\sneg^4$ is higher than that of $\sneg^3$, which suggests that our balance among $\sneg^3,\sneg^4,\sneg^5$ clearly improves the ability to identify the negative proposals, especially on the difficult proposals.

\subsection{BNPS compared with other sampling methods}
We have proved the rationality and effectiveness of our proposed BNPS in sampling ideal relationship pairs among a vast number of proposals. To further examine the superiority of BNPS, we implement several alternative sampling methods including \textit{online hard example mining (OHEM)} \cite{Shrivastava_2016_CVPR} and \textit{focal loss} \cite{Lin_2017_ICCV} and compare their capacities.

OHEM was first proposed in the object detection area and it prefers to sample harder examples than easier ones \cite{Shrivastava_2016_CVPR}. Typically, for a mini-batch of samples, only the top-$k$ most difficult proposals, i.e. proposals with top-$k$ highest loss values, will contribute to model optimization. In practice, the mini-batch is constructed to enforce a 1:3 ratio between positives and negatives, i.e. proposals in $S_{pos}$ and $S_{neg}$, to help ensure each mini-batch has enough positives. Since positive proposals are usually insufficient, we only apply OHEM on $S_{neg}$ to ensure this claim. We set mini-batch size $B\in\{48,64,96\}$ in our experiment.

The focal loss modifies the standard cross-entropy loss to dynamically down-weight the contribution of easy examples:
\begin{equation}
\begin{aligned}
    &\text{FL}(p_t) = -\alpha(1-p_t)^{\gamma}\log(p_t), \\
    &p_t =\left\{
             \begin{array}{ll}
            p & \text{if}\ y=1 \\
            1-p & \text{otherwise}.
             \end{array}
\right.
\end{aligned}
\end{equation}

In the above $y\in\{\pm 1\}$ specifies the ground-truth class and $p \in [0,1]$ is the model’s estimated probability for the class with label $y=1$. $\alpha$ and $\gamma$ are the balance coefficient and the tunable focusing parameter respectively. In our experiment, we use $\alpha=0.25$ with $\gamma\in\{1,2\}$.

Besides the above methods, Random Sampling and BNPS are also implemented as relationship pair sampling methods for comparison. We train models using these methods on V-COCO and keep all other settings consistent with setups in Section \ref{sec:setup}. Additionally, we assign $N=50$ for top-$N$ detections.

As results shown in Table \ref{tab:sampling_methods}, we observe that BNPS achieves the best performance, which reconfirms its superiority. Meanwhile, focal loss and OHEM in all settings behave worse than the basic random sampling method, which draws our attention. Focal loss and OHEM both emphasize the importance of samples with high loss values. However, in the scenario of relationship detection, these methods may focus too much on the hard samples and overlook the role of easy ones. Compared with them, our proposed BNPS uses the predefined sample classification method and pay equal attention to 5 classes of negative proposals, which makes our sampling strategy more robust towards the complicated distribution of samples in this scenario and less likely to be influenced by outliers.

\begin{table}[ht]
 	\centering
 	\setlength{\tabcolsep}{1.6mm}{
 	\begin{tabular}{ccc}
 	    \toprule
     	\multicolumn{1}{c}{Method} & Setup &
     	\multicolumn{1}{c}{AP$_{role}$} \\
        \hline
        \hline
        RS & - & 52.82 \\
        \hline
        \multirow{3}{*}{OHEM} & $B=48$ & 45.81 \\
        ~ & $B=64$ & 45.75 \\
        ~ & $B=96$ & 45.73 \\
        \hline
        \multirow{2}{*}{Focal Loss} & $\gamma=1$ & 52.55 \\
        ~ & $\gamma=2$ & 51.10 \\
        \hline
        BNPS & - & \textbf{54.69} \\
        
         \bottomrule
 	\end{tabular}}
 	\vspace{3mm}
 	\caption{Performance of the different relationship pair sampling methods. We use ResNet50-FPN as our backbone and modify the sampling strategy while keeping others unchanged.}
 	\label{tab:sampling_methods}
\end{table}
\end{document}